\documentclass{article}

\usepackage{PRIMEarxiv}

\usepackage[utf8]{inputenc} 
\usepackage[T1]{fontenc}    
\usepackage{hyperref}       
\usepackage{url}            
\usepackage{booktabs}       
\usepackage{amsfonts}       
\usepackage{nicefrac}       
\usepackage{microtype}      
\usepackage{lipsum}
\usepackage{fancyhdr}       
\usepackage{graphicx}       
\graphicspath{{media/}}     
\usepackage{amsmath}
\title{Integrating Implicit and Explicit Relational Biases through Graph-Based Multiple Instance Learning:\\A Case Study in Skin Lesion Diagnosis\thanks{Accepted as a short paper for presentation at the 21st International Conference on Computational Intelligence Methods for Bioinformatics and Biostatistics (CIBB 2026).}
}

\author{
Rafał Buler$^{1}$\thanks{Corresponding author: rafal.buler@pg.edu.pl} \and
Jakub Buler$^{1}$ \and
Maciej Bobowicz$^{2}$ \and
Michał Grochowski$^{1}$ \\[0.5em]
$^{1}$ Gdańsk University of Technology, Gdańsk, Poland \\
$^{2}$ Medical University of Gdańsk, Gdańsk, Poland \\
\texttt{\{rafal.buler, jakub.buler, michal.grochowski\}@pg.edu.pl} \\
\texttt{maciej.bobowicz@gumed.edu.pl}
}

\begin{document}
\maketitle

\begin{abstract}
Relational inductive biases are essential for capturing structural dependencies among data. This study investigates a dual-level relational framework for image classification, bridging the gap between implicit representation learning and explicit structural modelling. We begin by establishing a baseline using an EfficientNetB3 architecture. To move beyond standard convolutional biases, we adopt a patch-based strategy, employing a convolutional masked autoencoder to learn implicit inter-patch relationships through self-supervised reconstruction. We then extend this approach by incorporating explicit relational modelling, organizing the learned embeddings into various graph topologies, including grid-based, random, and k-nearest neighbour structures. Experimental results on the ISIC-2018 and ISIC-2019 skin lesion diagnosis benchmarks show that combining implicit inter-patch modelling with explicit graph-based message passing yields the best performance. On the ISIC-2018 test set, the baseline model achieves a balanced accuracy of 76.17\%, which improves to 77.12\% with implicit patch-based relational modelling. The fully integrated grid-structured Graph Attention Network further increases performance to 79.27\%. Similarly, on ISIC-2019, the implicit approach reaches 59.84\% balanced accuracy, while the combination of implicit and explicit modelling yields 60.67\%.
\end{abstract}


\section{Introduction}

Relational inductive biases guide how neural networks learn to capture dependencies within input data~\cite{battaglia2018relational}.
In the case of medical image analysis, feature extraction can be viewed as operating on a set of image entities, such as pixels, superpixels, or patches, with different modelling paradigms defining how relationships between these entities are captured. Deep convolutional neural networks treat individual pixels as fundamental entities and rely on hierarchical receptive fields to model spatial dependencies implicitly across the image. Superpixel-based approaches represent images as sets of homogeneous regions, where relationships between regions are modelled explicitly using graph neural networks (GNNs). Patch-based approaches instead represent images as collections of patch-level entities, which are typically encoded independently using convolutional networks, with relationships between patches introduced only at a downstream stage. As an alternative, Vision Transformer (ViT) architectures encode patches as tokens and model their interactions via self-attention~\cite{dosovitskiy2020image}.

The limited availability of annotated medical data has driven the adoption of self-supervised learning (SSL) for representation learning from unlabelled images. Early methods relied on pretext tasks to capture spatial and semantic structure, while more recent approaches are dominated by contrastive learning, which enforces invariance through data augmentation~\cite{gui2024survey}. However, such augmentations may distort clinically relevant features. Masked image modelling has therefore emerged as an alternative, learning representations by reconstructing masked regions and better preserving fine-grained structure~\cite{gao2022mcmae}.

In parallel, multiple instance learning (MIL) has become a standard framework for medical image classification when only image-level labels are available. Images are treated as bags of instances (e.g. patches), and attention-based pooling enables adaptive aggregation of informative regions~\cite{ilse2018attention}. More recently, graph-based extensions of MIL incorporate GNNs to explicitly model relationships between instances, refining patch representations through message passing~\cite{kipf2016semi}.

Despite these advances, it remains unclear whether relational modelling at the feature extraction stage alone is sufficient for downstream tasks, or whether combining it with explicit graph-based aggregation provides additional benefits. This study addresses this gap by analysing skin lesion classification, a task that requires integrating spatially distributed visual cues. 

Specifically, this work compares three complementary paradigms for medical image analysis. First, a conventional pipeline based on pretrained fully convolutional feature extraction applied to the entire image, followed by classification. Second, a self-supervised approach that learns patch-level representations through masked reconstruction, enabling the model to implicitly capture local spatial context without explicit relational modelling. Third, a two-stage strategy that augments such learned embeddings with graph-based aggregation, introducing explicit relational inductive bias between patches prior to classification. These approaches are evaluated on two datasets, with the aim of quantifying the impact of relational modelling at different stages.

\section{Methodology}
\begin{figure}[h]
    \centering
    \includegraphics[width=\linewidth]{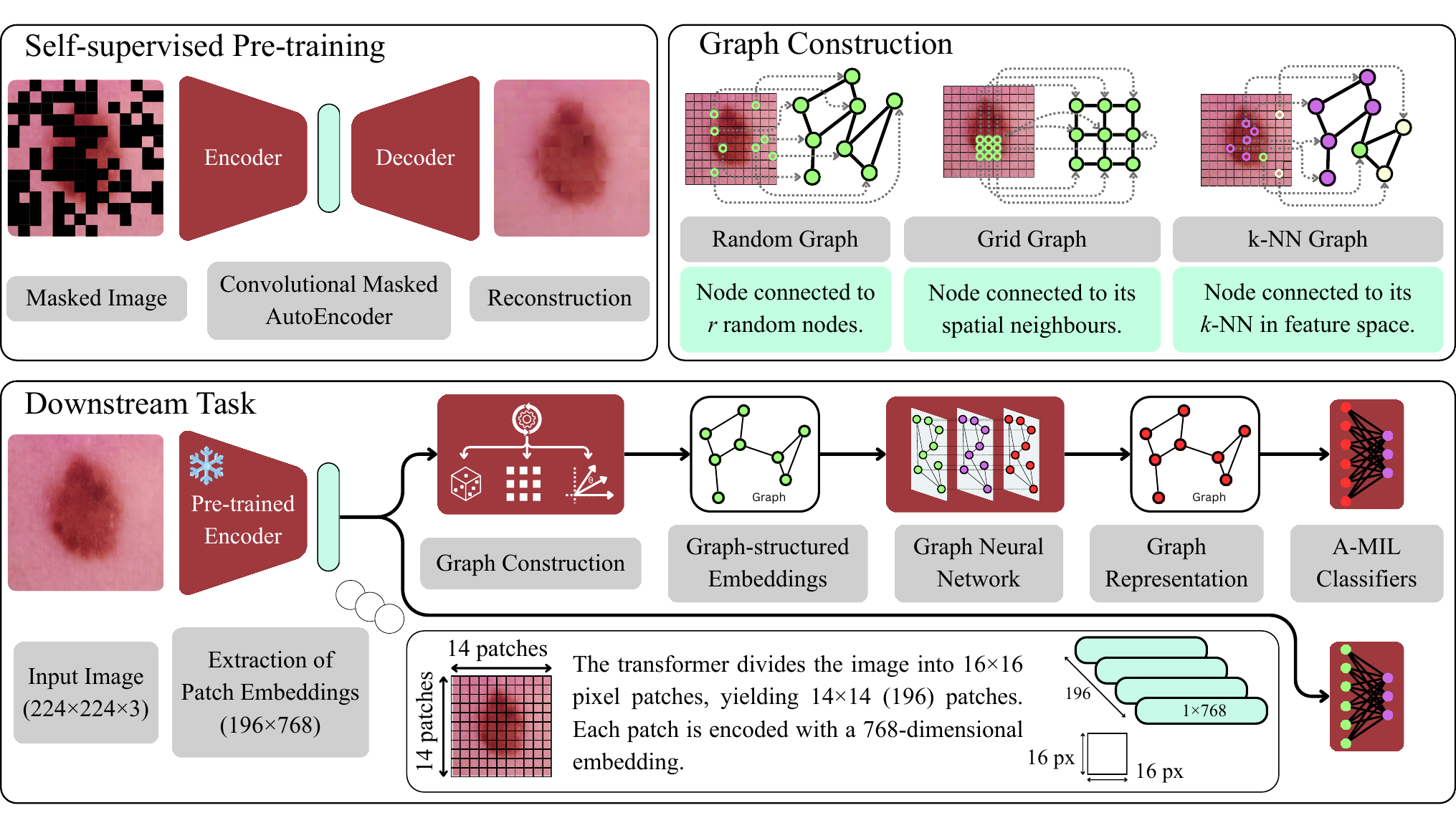}
    \caption{Pipeline overview: patch embeddings are aggregated using graph or non-graph MIL.}
    \label{fig:pipeline}
\end{figure}

The methodology instantiates this comparison within a unified pipeline. Patch-level representations are obtained from a frozen self-supervised autoencoder and used either directly in an attention-based MIL classifier or combined with graph-based aggregation via three construction strategies (random,  grid, kNN) to introduce explicit inter-patch interactions  (\autoref{fig:pipeline}).

\subsection{Dataset and Preprocessing}

The study utilised both the ISIC-2018 Challenge Task 3 dataset and the ISIC-2019 dataset, each providing dermoscopic images annotated with diagnostic labels and dedicated test sets. ISIC-2018 consists of 10{,}015 training and 1{,}514 test images across seven classes, while ISIC-2019 comprises 25{,}331 training and 8{,}238 test images, extending the taxonomy with one additional class and a further “unknown” category intended for out-of-distribution evaluation. The “unknown” samples were excluded from the experiments due to the absence of corresponding training data. For self-supervised representation learning, images from both datasets were combined and used jointly to train the masked autoencoder. As this stage does not rely on labels, no annotation information was used and therefore no label leakage occurs. To further avoid redundancy, duplicate images appearing across datasets were removed prior to pretraining. For downstream evaluation, the datasets were treated independently. On ISIC-2018, a five-fold cross-validation (CV) protocol was applied to the training set for model training and validation, while the official test set was used exclusively for final evaluation. Similarly, for ISIC-2019, models were trained on the training split and evaluated on its dedicated test set. This setup ensures a fair and consistent assessment on two independent benchmarks. All images were centre-cropped into a square, resized to $224 \times 224$~px, and normalised using ImageNet statistics. Class imbalance was addressed using weighted sampling during the training.

\subsection{Self-Supervised Patch Representation Learning}

Patch-level representations were learned using a convolutional masked autoencoder with a ViT backbone (ConvMAE)~\cite{gao2022mcmae}. During pre-training, several configurations were explored by different masking ratios (0.25–0.75) and reconstruction losses, including mean squared error and its normalized variant computed after per-patch normalization of target pixels. The model selection was based on linear probing performance on ISIC-2018 after discarding the decoder and freezing the encoder. As SSL is used here to obtain representations rather than optimise pretraining performance, this comparison is kept minimal to focus on downstream relational modelling. The selected encoder was then fixed and used across all experiments to ensure a consistent feature space. Each image was decomposed into $14 \times 14 = 196$ non-overlapping patches (size $16 \times 16$ each), producing embeddings $\mathbf{x}_i$. This patch-level representation naturally supports a MIL formulation used in subsequent experiments.
\subsection{Multiple Instance Learning Baseline}

As a baseline, image-level classification was performed using an attention-based multiple instance learning (AMIL) model adopted from~\cite{ilse2018attention}.
Each image was represented as a bag of patches, where the patch embeddings produced by the encoder were treated as independent instances. Formally, let an image be represented by a bag
$
\mathcal{X} = \{ \mathbf{x}_i \}_{i=1}^{N}, \quad \mathbf{x}_i \in \mathbb{R}^{d},
$
where $N = 196$ denotes the number of patches and $d = 768$ is the dimensionality of each patch embedding, corresponding to the output channel size of the final transformer block, which balances representation capacity and computational efficiency consistent with standard MAE and ViT architectures.
The bag $\mathcal{X}$ corresponds to a single image-level label $y$, while no supervision is available at the instance-level. An attention mechanism was used to compute a normalised importance weight $a_i$ for each instance $\mathbf{x}_i$:
\begin{equation}
a_i = \frac{\exp\left( \mathbf{w}^\top \tanh(\mathbf{V}\mathbf{x}_i) \right)}{\sum_{j=1}^{N} \exp\left( \mathbf{w}^\top \tanh(\mathbf{V}\mathbf{x}_j) \right)},
\end{equation}
where $\mathbf{V}$ and $\mathbf{w}$ are learnable parameters.
The image-level representation was then obtained by pooling the patch embeddings using an attention mechanism:
\begin{equation}
\mathbf{z} = \sum_{i=1}^{N} a_i \mathbf{x}_i.
\end{equation}

The aggregated representation $\mathbf{z}$ was subsequently passed to a classifier to produce the final image-level prediction. This MIL formulation is permutation-invariant with respect to patch ordering and treats all patch embeddings as independent instances, without explicitly modelling spatial or semantic relationships. It therefore serves as a reference point for assessing the effect of introducing explicit inter-patch relational modelling in subsequent variants.

\subsection{Graph Construction over Patch Representations}

To incorporate the relational structure among the patch embeddings, the set of patch-level features was reformulated as an undirected graph.
Specifically, each image was represented as a graph
$\mathcal{G} = (\mathcal{V}, \mathcal{E})$, 
where the node set $\mathcal{V} = \{ v_i \}_{i=1}^{N}$ corresponded to the patch embeddings, and each node $v_i$ was associated with the feature vector $\mathbf{x}_i \in \mathbb{R}^{d}$.
The edge set $\mathcal{E}$ defines the relationships between patches and was constructed according to different strategies.

The random graph serves as a non-structured baseline, where each node $v_i$ is connected to up to $r$ randomly sampled nodes from $\mathcal{V}$, with $r \in {1,2,3,4}$. This spans from a sparsely connected setting close to the MIL case to a denser configuration comparable to a grid graph, allowing isolation of connectivity effects from structured inductive biases.
The MIL baseline corresponds to the degenerate case $\mathcal{E} = \emptyset$, where no message passing is performed and patch embeddings are aggregated directly via attention pooling without explicit relational modelling.
The grid graph is defined using the original $14 \times 14$ spatial layout, where edges connect spatially adjacent patches. An edge $(v_i, v_j) \in \mathcal{E}$ is added if the corresponding patches are neighbours on the grid. Two variants are considered: 4-neighbourhood connectivity and 8-neighbourhood connectivity.
The k-nearest neighbour (kNN) graph defines edges in feature space based on Euclidean distance. For each node $v_i$, edges are added to its $k$ nearest neighbours:
\begin{equation}
\mathcal{E} = { (v_i, v_j) \mid v_j \in \text{kNN}(\mathbf{x}_i) }.
\end{equation}

\subsection{Graph Neural Networks with MIL Pooling}
Graph-structured patch representations were processed with GNNs to model inter-patch interactions prior to aggregation. This included multiple variants to be analysed: graph convolutional networks (GCN), graph attention networks (GAT), and graph transformer-based architectures. Given a graph $\mathcal{G} = (\mathcal{V}, \mathcal{E})$ with node features $\{ \mathbf{x}_i \}_{i=1}^{N}$, each layer performs message passing: \begin{equation} \mathbf{h}_i^{(l+1)} = \phi \left( \mathbf{h}_i^{(l)}, { \mathbf{h}_j^{(l)} \mid (v_i, v_j) \in \mathcal{E} } \right), \end{equation} where $\mathbf{h}_i^{(0)} = \mathbf{x}_i$ and $\phi(\cdot)$ is a learnable aggregation function. After $L$ layers, node embeddings ${ \mathbf{h}_i^{(L)} }$ were aggregated using the same attention-based MIL pooling as in the baseline, followed by classification. This ensures that differences relative to the MIL baseline arise solely from relational modelling via message passing. To isolate the effect of relational learning from increased model capacity, an additional setting with frozen GNN layers was evaluated. Here, message passing was performed with fixed weights and only the MIL classifier was trained.

\subsection{Hyperparameter Tuning Protocol}

Hyperparameter optimisation was conducted under a unified protocol across all models, including the MIL baseline and graph-based variants. Shared hyperparameters included the learning rate, optimiser type, dropout rate, and the hidden dimensions of both the classifier and attention layers. For graph-based models, additional parameters were tuned, including the GNN type, number of message-passing layers, hidden dimensions, graph construction strategy, diagonal connectivity in grid graphs, and neighbourhood size in kNN and random graphs. Tuning was performed using five-fold CV on the training sets, with balanced accuracy as the selection metric. Reported results correspond to the best-performing hyperparameter configurations.

\section{Results}
To establish a strong foundation for the downstream task, patch embeddings were first learnt using a ConvMAE. Linear probing of the ConvMAE encoder on the ISIC-2018 training set showed that a masking ratio of $0.50$ resulted in the lowest reconstruction loss, indicating optimal patch reconstruction. Therefore, the encoder trained on $0.50$ masking ratio was selected for feature extraction in all downstream task based on AMIL and Graph AMIL based aggregation.

Building on SSL-based patch representations, AMIL was used for image-level aggregation, capturing relational information both implicitly in the embedding space and explicitly via graph-based message passing. The AMIL model achieved a mean balanced accuracy (BAcc) of $77.12\%$, outperforming a conventional end-to-end CNN baseline ($76.17\%$). Introducing explicit relational modelling further improved performance. Random graphs yielded moderate gains (best at $r=2$), while the highest BAcc of $79.27\%$ was obtained with a grid graph without diagonal connections combined with GAT layers. The freezing of GNN parameters resulted in a considerable decrease, suggesting that learned relational modelling, rather than augmented capacity, is a primary factor contributing to performance. Overall, grid-based MIL outperforms prior single-model approaches and remains competitive with ensemble methods that aggregate predictions from up to 90 deep neural models. On ISIC-2019, the proposed approach achieves strong performance (60.67\% BAcc) with a kNN graph ($k=4$) processed with GAT layers, surpassing several existing methods and remaining competitive with more complex ensemble-based solutions. All evaluated approaches are reported in~\autoref{tab:results_summary}.

\begin{table*}[!htb]
\centering
\scriptsize
\caption{Performance comparison on ISIC-2018 (left) and ISIC-2019 (right) dedicated test sets. Results are reported as mean ± SD over 5-fold cross-validation. \textbf{Results from prior work are included only when explicitly reported on the corresponding dedicated test set.} In the table, \textit{P} denotes pixel-level interactions, \textit{I} implicit patch relational modelling, and \textit{E} explicit patch relational modelling. $^{\dagger}$Results obtained from the official \href{https://challenge.isic-archive.com/leaderboards/2019/?task=52&group_by=team}{ISIC-2019 challenge leaderboard}.
}
\label{tab:results_summary}
\begin{minipage}[t]{0.45\linewidth}
\centering
\vspace{0pt}

\begin{tabular}{llc}
\toprule
Model & Relational & BAcc [\%] \\
\midrule
EfficientNet-B3 & $P$ & 76.17 $\pm$ 0.89 \\
Base AMIL & $P+I$ & 77.12 $\pm$ 1.49 \\
Graph AMIL (r=2) & $P+I+E_{Random}$ & 78.24 $\pm$ 0.86 \\
Graph AMIL (Frozen) & $P+I+E_{Grid}$ & 34.28 ± 3.39\\
Graph AMIL & $P+I+E_{Grid}$ & \textbf{79.27 $\pm$ 1.38} \\
Deep Hierarchical~\cite{barata2021explainable} & - & 72.60\\
Ensemble (4 models)~\cite{lee2018wonderm} & - & 78.50\\
Ensemble (25 models)~\cite{tabibi2024ensemble} & - & 84.00\\
Ensemble (90 models)~\cite{mahbod2020transfer} & - & \textbf{86.20}\\
\bottomrule
\end{tabular}
\end{minipage}
\hfill
\begin{minipage}[t]{0.45\linewidth}
\centering
\vspace{0pt}
\begin{tabular}{llc}
\toprule
Model & Relational & BAcc [\%] \\
\midrule
Base AMIL                  & $P+I$          & 59.84 $\pm$ 1.50  \\

Graph AMIL                 & $P+I+E_{kNN_4}$  & \textbf{60.67 $\pm$ 0.68} \\
Non-Ensemble-Top-5$^{\dagger}$ & -              & 56.90 \\
Ensemble-Top-4$^{\dagger}$ & -              & 57.80 \\
Ensemble-Top-3$^{\dagger}$ & -              & 59.30 \\
Ensemble-Top-2$^{\dagger}$ & -              & 60.70 \\
Ensemble-Top-1$^{\dagger}$ & -              & \textbf{63.60} \\
\bottomrule
\end{tabular}
\end{minipage}
\end{table*}

\section{Discussion and Conclusion}

Our results demonstrate that the self-supervised patch representations learnt via masked image modelling provide a robust foundation for downstream classification. Incorporating explicit relational modelling through GNNs further improves performance. The substantial drop in performance when freezing the GNN parameters underscores the importance of learnt message passing for capturing meaningful inter-patch relationships. These findings indicate that relational information can be effectively encoded in an intermediate manner during the SSL and enhanced through additional explicit graph modelling. The proposed relationally aware framework achieves competitive results compared to previous single-model deep learning approaches and ensemble methods, highlighting the value of combining self-supervised learning with graph-based relational modelling in image analysis.
This study emphasises the complementary roles of implicit and explicit relational modelling in patch-based classification, providing guidance for the design of efficient and effective architectures for image analysis. The proposed approach exhibits strong generalisation capabilities, demonstrating that it can be applied to model data relations and solve diagnostic and classification tasks across other medical domains.

\section*{Conflict of interests}
\label{sec:CONFLICT-OF-INTERESTS}
\noindent The authors declare that they have no competing interests.

\section*{Acknowledgments}
\label{sec:ACKNOWLEDGMENTS}
\noindent The authors acknowledge the EUCAIM \#101100633 consortium and the Centre of Informatics Tricity Academic Supercomputer and networK CI-TASK (project no. 01191) for their support. 

\section*{Availability of data and software code}
\label{sec:AVAILABILITY}
\noindent The datasets used in this study are publicly available. The software code is available from the corresponding author upon reasonable request.

\bibliographystyle{unsrt}  
\bibliography{references}

\end{document}